\documentclass[runningheads]{llncs}
\usepackage[T1]{fontenc}
\usepackage{graphicx}
\usepackage{hyperref}
\usepackage{color}

\usepackage{xspace}
\usepackage[inline]{enumitem}
\usepackage{booktabs}
\usepackage{makecell}
\usepackage{multicol}
\usepackage{multirow}
\newcommand{\mtg}{\emph{Magic: the Gathering}\xspace}
\newcommand{\hs}{\emph{Hearthstone}\xspace}
\newcommand{\pokemon}{\emph{Pokémon TCG}\xspace} 
\newcommand{\loru}{\emph{Legends of Runeterra}\xspace}
\newcommand{\fab}{\emph{Flesh and Blood}\xspace}
\newcommand{\yugioh}{\emph{Yu-Gi-Oh! TCG}\xspace}
\newcommand{\locm}{\emph{Legends of Code and Magic}\xspace}
\newcommand{\gwt}{\emph{Gwent}\xspace}
\newcommand{\ms}{\emph{Marvel Snap}\xspace}
\newcommand{\tot}{\emph{Tales of Tribute}\xspace}
\begin{document}
\title{A Taxonomy of Collectible Card Games \\from a Game-Playing AI Perspective}
\titlerunning{A Taxonomy of CCGs from a Game-Playing AI Perspective}
%
\author{Ronaldo e Silva Vieira\inst{1}\orcidID{0000-0002-7109-0897} \and
Anderson Rocha Tavares\inst{2}\orcidID{0000-0002-8530-6468} \and
Luiz Chaimowicz\inst{1}\orcidID{0000-0001-8156-9941}}
\authorrunning{R. e Silva Vieira, A. Rocha Tavares, and L. Chaimowicz}
%
\institute{Universidade Federal de Minas Gerais, Belo Horizonte, Brazil\\
\email{\{ronaldo.vieira,chaimo\}@dcc.ufmg.br} \and
Universidade Federal do Rio Grande do Sul, Porto Alegre, Brazil
\email{artavares@inf.ufrgs.br}}
\maketitle              
\begin{abstract}
Collectible card games are challenging, widely played games that have received increasing attention from the AI research community in recent years. 
Despite important breakthroughs, the field still poses many unresolved challenges.
This work aims to help further research on the genre by proposing a taxonomy of collectible card games by analyzing their rules, mechanics, and game modes from the perspective of game-playing AI research.
To achieve this, we studied a set of popular games and provided a thorough discussion about their characteristics.
\end{abstract}
\section{Introduction}
\label{sec:introduction}

Collectible card games (CCGs), also known as trading card games, are notoriously complex games with steep learning curves for human players. As of the time of writing, the  \mtg comprehensive rules book contains 290 pages.\footnote{\url{https://magic.wizards.com/en/rules}}
Human players are not alone: CCGs are also difficult for computers, as their complexity also have a computational aspect. For example, the same \mtg is Turing complete \cite{ChurchillBH21}, and even determining whether a specific action type is valid is coNP-complete \cite{ChatterjeeI16}. 

Nonetheless, this does not prevent computers from playing CCGs well. The last decade has seen a rise in the popularity of  AI for CCGs. Building on the success in less complex games such as Go \cite{SilverSSAHGHBLB17} and Texas Hold'em Poker \cite{science.aay2400}, in recent years, a deep reinforcement learning AI agent beat a top-level human player three wins to none in a best-of-five match of \hs and is regarded as the first computer player to achieve such a feat \cite{XiaoZHHCS23}. 

Despite the recent progress, the investigation of AI for CCGs is still a very open field. Playing the game to win is just one of the many interesting applications of AI in CCGs \cite{HooverTLS20}, and, even so, there are still many open challenges, such as leveraging computer players for playtesting, scaling up current methodologies to more complex CCGs, and building game-agnostic approaches.

This paper aims to support further research on CCGs by tackling the following research question: \textbf{what are the differences and similarities among the rules, mechanics, and game modes of popular collectible card games?}
To do so, we study a set of CCGs and propose a taxonomy of the genre, focusing on aspects relevant to game-playing AI.
We do not intend to cover every existing CCG or every aspect of CCGs, as it is not feasible.\footnote{As of the time of writing, the ``list of collectible card games'' on Wikipedia has 391 physical CCGs \cite{list-of-ccgs-wikipedia}. Steam's trading card game category has 553 digital CCGs \cite{list-of-ccgs-steam}.} We do, however, aim to be representative.
We believe this paper benefits researchers and enthusiasts in the game-playing AI field who are entering the CCG genre, as well as those in the CCG genre who are entering game-playing AI.
It may also benefit indie CCG and game designers to improve and create new games.

\section{Related Work}
\label{sec:rel-work}

To the best of our knowledge, no other works propose taxonomies of CCGs.
The work of Hoover et. al \cite{HooverTLS20}, however, discusses the many AI problems of \hs. 
Although \hs shares many characteristics with other CCGs, our analysis of a broader set of games enabled us to propose a taxonomy and discuss nuances among them.
Moreover, we adopt the perspective of game-playing AI, while Hoover et al.'s work also considers areas such as content generation, player modeling, and game balancing.

In the past, competitions held at academic conferences were an important force behind the rise in popularity of AI for CCGs \cite{JanuszTS17,Dockhorn2019,Kowalski2023,totai-competition}. 
So far, there has been research in battles \cite{CowlingWP12,byterl,XiaoZHHCS23} as well as in all deck-building modes as we discuss in Section \ref{sec:analysis}:
constructed \cite{bjorke2017deckbuilding,BhattLSWTH2018,Garcia-SanchezT18,creativestone}, independent draft \cite{kowalski2020,YangYC21,VieiraTC23}, and round-robin draft \cite{WardMBTK21,BertramFM21}. 
Work on the computational complexity of playing CCGs has also been done. Churchill et. al \cite{ChurchillBH21} proved that finding the optimal play in \mtg is undecidable, appointing it as ``the most computationally complex real-world game known in the literature''. Another work demonstrated that optimal play is as hard as arithmetic \cite{abs-2003-05119}. Furthermore, checking the legality of a particular type of action in \textit{Magic} is found to be coNP-complete \cite{ChatterjeeI16}.
Regarding \hs battles, the mate-in-$n$ problem (whether the player can win in $n$ turns) 
in a simpler variant of the game has been found to be PSPACE-hard \cite{abs-2305-12731}, and mate-in-1 in specific game modes to be NP-hard \cite{HoffmannLW20}. We did not find such analyses regarding other CCGs.

\section{Scope}
\label{sec:data-collection}

To answer our research question, we first reviewed the rules, mechanics, and game modes of different CCGs. We considered games that are currently active, are available in English, and contain both strategic deck-building and battles as completely separate processes or phases, a genre mostly inspired by the pioneer game \mtg.
From all compatible CCGs, we selected 10 representative games, considering their popularity, relevance to AI research, and diversity of rules. 
The selected games were: 
\fab, \gwt,\footnote{The standalone \gwt, not the version present in \textit{The Witcher 3: Wild Hunt} \cite{the-witcher-3}.} \hs, \locm, \loru, \mtg, \ms, \pokemon, \tot,\footnote{\tot does not include a deck building process, but we consider it 
in our study as it is currently used in the \textit{Tales of Tribute AI competition} \cite{totai-competition}, held at the IEEE Conference on Games.} and \yugioh. 
To study them, we referred to, in order of priority:
    the game's official rules book or web page 
    \cite{fab-rulesbook,fab-website,locm-website,mtg-rulesbook,pokemontcg-rulesbook,yugiohtcg-rulesbook},
    \textit{wikis} and other player-managed knowledge bases 
    \cite{pk-wiki,tot-wiki,tot-wiki2,hs-wiki,hearthsim,gwt-wiki,lor-wiki,ms-wiki,marvelsnapzone,mtg-wiki},
    how-to-play tutorials 
    \cite{tot-tutorial,tot-tutorial2,ms-tutorial,lor-tutorial},
    related entries in Q\&A platforms, and
    our own in-game experience.

Our study did not consider commercial aspects (e.g., distribution model, booster statistics, card economy), collectible aspects (e.g., card cosmetics and lore, card crafting), competitive aspects (e.g., professional play, metagame analysis), rules and mechanics not directly related to CCG gameplay (e.g., story mode, season pass), temporary or experimental game modes (e.g., weekly rotating game modes), and game modes with more than two players per battle.
A comprehensive summary of the results is 
available online\footnote{Available at \url{https://bit.ly/4cFdz0v}.}
and served as a foundation for our answers to our research question.

\section{Taxonomy}
\label{sec:analysis}

In this section, we describe the general structure of the reviewed CCGs and discuss similarities and differences regarding rules, mechanics, and game modes.
In general, CCGs use different terms for the same concepts. 
Thus, in the remainder of this paper, we select the term we deem more appropriate whenever there is more than one alternative. We use \textbf{bold} for key term definitions and \textit{italic} for names of games, secondary term definitions, or emphasis.

From an AI standpoint, the act of playing CCGs can be divided into two primary, ordered parts: deck-building and battling. 
In the \textbf{deck-building part}, players are tasked to build a deck. 
While traditional card games use the standard 52-card deck, CCGs have their own card designs (which we discuss in Section \ref{sec:cards}), and players should select a subset of cards among the (usually very large) set of available cards to build their own deck.
This \textit{card pool} contains diverse cards that enable many different deck strategies. Having multiple copies of the same card in a deck is usually allowed but limited. It is desirable that the cards in a deck have some degree of synergy, that is, be resonant with each other in terms of strategy. 
The specific rules and the process of deck-building are specified by the game \textit{mode} being played (Section \ref{sec:modes}), while the card pool is determined by the game \textit{format} being played.\footnote{The terms \textit{mode} and \textit{format} are sometimes used interchangeably. We adhere to the following definition: a mode specifies the deck-building rules, and a format specifies the card pool. We do not discuss formats since they are primarily commercial.} In some CCGs, players must also choose a deck \textit{class} (Section \ref{sec:classes}), which further limits the card pool.

After building a deck, its owner uses it to battle other players. The \textbf{battle part} is often considered the actual gameplay of a CCG, where the player usually impersonates or represents a \textit{hero} (Section \ref{sec:classes}), invoking creatures and casting spells in order to defeat their opponent.
At the start of a battle, both players have their decks shuffled and positioned face down. They draw a predetermined number of cards as their starting hand, and some games allow partial or total hand redraw (Section \ref{sec:mulligan}). Then, players take turns (Section \ref{sec:turns}) in which they draw cards, and manage resources (Section \ref{sec:resource-gen}) to perform actions (Section \ref{sec:actions}). Actions involve playing cards, which belong to a game zone (Section \ref{sec:zones}), activating card abilities, and handling combats (Section \ref{sec:combat}). 
The battle goes on until a win condition (Section \ref{sec:win-con}) is reached.

\subsection{Cards}
\label{sec:cards}

The card is, unsurprisingly, the most fundamental concept of CCGs. Almost every piece of information in the game concerns a card, and almost every decision in the game stems from or targets a card.
Cards have a name, an artwork, and sometimes a \textit{flavor text}, all not directly relevant to gameplay but helpful to situate that card within the game's lore and stimulate collection.
In addition, cards have at least a \emph{type}, some \emph{attributes}, and may have a \emph{class} and \emph{abilities}, all of which are gameplay-related.
Different CCGs have different visual identities and, therefore, organize these pieces of information differently. 

The \textit{class} of a card imposes direct or indirect restrictions on which cards can be in the same deck. In CCGs with a class system, cards normally belong to one class or are neutral, and can exceptionally belong to multiple classes. The \textit{attributes} of a card, in turn, are numeric variables related to resource management and the game's main win conditions; for instance, resource cost, attack, and health attributes. A card's \textit{type} categorizes it and specifies its high-level rules: which attributes it has, when and how it should be played, whether it stays on the battlefield, what happens when it is destroyed, etc.
We have identified seven established card types among the CCGs we studied:
\textit{creature}, \textit{resource}, \textit{spell}, \textit{enchantment}, \textit{equipment}, \textit{location}, and \textit{toolbox}.

\textbf{Creatures} are central to winning. In combat-based CCGs (as we define in Section \ref{sec:win-con}), they have attack and health attributes and can attack and be attacked, whereas, in conquest-based CCGs, they provide influence points as long as they are on the battlefield. \textbf{Resource cards}, on the battlefield, can generate resource points once per turn. The most basic resource cards usually generate a single resource point per turn and are allowed unlimited copies in a deck. \textbf{Equipment cards} are bound to a creature and provide local effects (i.e., they modify that creature's attributes and abilities), and \textbf{location cards}, in turn, provide global effects (i.e, they modify many cards at once). \textbf{Spell cards} can provide local or global effects and differ from other types because they are destroyed immediately after being played and, consequently, do not stay on the battlefield. Finally, \textbf{toolbox cards} provide a set of activated abilities at the expense of their health points.

Lastly, a card can have many abilities. These abilities can be expressed in natural language or as a \textit{keyword} (a frequently occurring ability whose interpretation is provided for in the game's rules book). They can be of three kinds: \textit{triggered}, \textit{activated}, or \textit{static}. 
\textbf{Triggered abilities} and \textbf{activated abilities} have \textit{prerequisites} and \textit{effects}: whenever their prerequisites are fulfilled, they yield effects, altering the state of the battle or the rules of the game.
They differ in that the effects of triggered abilities occur whenever specified conditions are met, while the effects of activated abilities occur whenever the player activates them by paying a specified cost (in resource points and/or in some other effect that is detrimental to the player). Unless specified otherwise, triggered and activated abilities have no limit of use.
\textbf{Static abilities}, in contrast, have no prerequisites and thus apply their effects continuously for the duration that the card remains on the battlefield (or at all times, in some CCGs).

\subsection{Game Modes}
\label{sec:modes}

Most CCGs present various game modes, that is, different ways of playing the game, each with a distinct set of rules, mostly regarding deck-building.
Across the games we analyzed, we recognize three main game modes: constructed, independent draft, and round-robin draft. 

\textbf{Constructed} is the simplest and the most common form of play. Players choose a subset of cards from a predetermined, large set of cards (the card pool) to build a deck. The minimum size of a deck is enforced, and sometimes a maximum size is enforced as well. CCGs often allow 2 to 4 copies of a single card 
and unlimited copies of any basic resource card. There may be additional card copy restrictions regarding some card types or rarities. A ``sideboard'' deck with up to 15 cards may be allowed when playing multiple battles; cards can be swapped or added to the main deck between battles.

We further subdivide constructed deck-building into its offline and online forms. In \textbf{offline} constructed, the card pool is known beforehand. 
On the other hand, in \textbf{online} constructed, the card pool is known only at the time of play and is often a random sample of an offline card pool. Players are given a fixed amount of time to build their decks, and there are no card copy restrictions. From the games we analyzed, \locm is the only one to procedurally generate the card pool in its online constructed mode.

The other game modes we discuss are draft-based modes, in the sense that players build their decks via a \textit{drafting} process: each turn, players receive a small, random sample of cards and pick one among them; this goes on for a fixed number of turns. In either draft-based variant, there are no card copy restrictions.
In the \textbf{independent draft} variant, the samples of cards may (\locm) or may not (\gwt, \hs, and \loru) be the same for all players. 
In \gwt and \hs, a deck class must also be drafted. After the drafting process, each player's picked cards become their deck, and they battle until they reach a certain number of wins or 3 losses, receiving in-game rewards based on the number of wins. This game mode is appropriate for digital CCGs, as players need not build their decks simultaneously or together.

On the other hand, in the \textbf{round-robin draft} variant, the participant players build their decks simultaneously. The initial sample of cards is usually the content of a booster pack.\footnote{A \textit{booster} is a pack of random cards sold in many CCGs.} After a player chooses a card, the remaining cards are passed to the next player and become its sample in the next turn (hence, a \textit{round-robin} draft). This happens until there are no remaining cards. In \fab and \mtg, a total of three booster packs are used in this process, and the first and third are passed to the player on the left, while the second is passed to the player on the right.
After this process, each player's chosen cards become a personal card pool from which they must build a deck (basic resource cards can be added as desired). In \pokemon, four booster packs are used, and the personal card pool has a premade deck in addition.

\subsection{Classes and Heroes}
\label{sec:classes}

CCGs may have a deck class system. 
We differentiate two variants of the implementation of deck classes in CCGs. \textbf{Hard-classed} games force the deck to contain only cards of the chosen class or neutral cards. This is true for \fab, \gwt, \hs, and \tot. The first and the latter also present decks with multiple classes. \textbf{Soft-classed} games allow deck-building with cards of all classes; however, in general, it is strategically convenient to restrict the deck to cards of few classes or a single class. This includes the colors (red, blue, etc.) in \mtg and the types (water, grass, fire, etc.) in \pokemon and \yugioh. \locm and \ms have no class system.

In CCGs, there is usually the figure of a \textbf{hero}: a role the player represents; one that summons creatures, uses spells and equipment to battle their opponent, another hero. The hero can be an implicit character such as in \mtg, where the player is referred to as ``planeswalker'', and in \pokemon as the Pokémon trainer; or can be explicit like the hero in \fab and \hs and the leader in \gwt, which are in-game entities. In some games, an (explicit) hero must be chosen for a deck and limits which cards are available for deck-building---acting as the CCG's class system. 

In \loru, the concept of heroes is different: they are a card type akin to creatures, and multiple heroes can be present in a deck. In \mtg, similarly, 
the so-called planeswalkers are also a type of toolbox card, 
despite the players themselves being the implicit heroes. In \gwt, \hs, and \tot, explicit heroes act as indestructible creatures or toolbox cards. In \fab, the hero is, semantically, the main actor in combat instead of creatures, and the equivalent to creatures cards are tools, weapons, spells, or actions that the hero may use. In the remaining CCGs we analyzed, the hero is implicit.

\subsection{Redraw}
\label{sec:mulligan}

At the start of a battle, after the players draw their starting cards, many CCGs offer a chance to redraw them. 
This is convenient when a starting hand does not appear competitive.
This mechanism is popularly known as ``mulligan'' \cite{mulligan}.

When games offer this mechanism, it is either a \textbf{card redraw}, in which the player selects specific cards from their starting hand to be redrawn, or a \textbf{hand redraw}, in which the whole hand is redrawn with some form of penalty: receiving one less card than before or having the opponent receive one more card. In both variants, the selected cards return to the deck, which is shuffled again before redraw.
In \textit{Pokémon: TCG}, hand redraw is compulsory whenever the starting hand contains no creatures. 
\mtg, in turn, has seen different rulings about hand redrawing throughout its history. The current one is called \textit{London Mulligan} and consists of redrawing the hand but, afterwards, choosing one card to be put back at the bottom of the deck for each time the player has redrawn \cite{duke_2019}.
In the games we analyzed, hand redraw may be performed multiple times (with cumulative penalties), whereas card redraw may not.

\subsection{Turns}
\label{sec:turns}
After deciding whether to redraw (if applicable), players start taking turns. These turns have a predetermined structure, and while they vary across CCGs, there is a general structure composed of phases. We define phases as the different parts of a turn where the valid actions differ or where key events occur.

Turns usually start with a \textbf{draw phase}, where the active player draws cards (usually one) from their deck to their hand. 
\fab is an exception to this case, where this happens at the end of the turn, and the player draws cards up to their hero's ``intellect'' attribute.
Then, the \textbf{main phase} starts, in which the active player can perform most of the actions, such as playing cards, activating abilities, and attacking. In some combat-based CCGs, however, the \textbf{combat phase} (described in Section \ref{sec:combat}) is the moment where attacking is allowed. There may be an additional main phase after combat, and, sometimes, an \textbf{end phase} reserved for abilities that trigger or terminate ``at the end of the turn''.\looseness=-1

In most CCGs, players can \textit{respond} to some of the opponent's actions and even their own. Usually, just a subset of cards can be played (called \textit{fast}, \textit{instant}, or \textit{reaction} cards), and a subset of other actions can be performed as a response. 
We can regard this as a different phase, a \textbf{response phase}. Moreover, players can also respond to responses. Most CCGs choose to resolve a chain of responses in a last-in-first-out manner (forming a \textit{stack}) in which responses are executed before the responded action. \hs uses a first-in-first-out manner (forming a \textit{queue} instead). These chains of responses enable mechanics such as ``countering'', in which, for example, a card's effect may be to annul an action currently in the stack or queue. \loru limits its stack to at most 10 cards (original action plus 9 responses).

In most CCGs, turns are carried out \textbf{alternately}, with only one player (the active player) primarily performing actions until the turn is passed to the other player.
However, in \loru, turns are \textbf{joint}. In other words, both players draw a card and can perform actions. In this case, after a player performs an action, priority is given to the opponent to act. The concept of an active player still exists, indicating the player that receives priority first and who is allowed to attack.
Finally, in \ms, turns are \textbf{simultaneous}: both players draw at the same time and are allowed to play cards face down. When both players finish or time runs out, the cards are revealed, and their effects are resolved.

\subsection{Resource Systems}
\label{sec:resource-gen}

Most CCG battles revolve around a single primary resource, usually called \textit{mana}, required for playing cards and sometimes to activate abilities. The mana cost of a card correlates positively with its overall power. Some CCGs have different types of mana, and cards or abilities may require specific mana types.
In the CCGs we reviewed, mana is obtained in one of three ways, in order of complexity: incrementally, from resource cards, or from discarding cards.\looseness=-1

In \hs, \locm, \loru, and \ms, mana is obtained \textbf{incrementally} as the battle progresses. At the player's first turn, one mana point is available, 
and so on: at the $k$-th turn, $k$ mana points are available, up until a limit (usually 10 or 12).
In \mtg, \pokemon, and \tot, mana comes \textbf{from resource cards}. Players can play one resource card per turn, and most generate one mana point per turn, which makes progression similar to the incremental mechanic. A key difference is that, when building a deck, players must balance the ratio of resource cards to other cards, as well as the mana types they generate or require.
In \fab, mana is generated \textbf{by discarding cards}. Every card in the game has three versions, which generate one, two, or three mana points, with the card's power varying slightly, inversely proportional to the mana they generate. Besides, destroyed and discarded cards during battle are placed at the bottom of their owner's deck. This resource system raises a contrasting challenge in deck-building compared to the others.

Lastly, \gwt and \yugioh have no resource similar to mana. In the former, card draw is severely limited, so while cards have no cost to be played, creatures' position on the battlefield and order of play are determinant. In the latter, creature cards on the battlefield are used ``as sacrifice'' for more powerful creatures to be played.

Unspent mana usually disappears at the end of the turn or phase. In \loru, up to three unspent mana points are stored indefinitely as ``spell mana'' for playing spell cards. 
In general, optimal play is related to optimal mana use---i.e., efficiently converting mana to creatures, spells, or abilities and using those to win.

\subsection{Zones}
\label{sec:zones}

The different sets of cards present in a battle, which are usually symmetric (each player has their own) and have different rules regarding the cards in it, are called zones.
A genre-defining characteristic of CCGs is the presence of a \textbf{deck}, \textbf{hand}, and \textbf{battlefield} zone for each player.
All CCGs have one or more battlefield zones, which can be further divided into \textbf{lanes} that may be identical or very similar (in terms of rules). In most CCGs, when battlefield cards are destroyed, they go to a \textbf{graveyard} zone. Some effects may remove cards from the game, moving them to a ``\textbf{outside the game}'' zone. The \textbf{stack} or \textbf{queue} is also a zone, although, in some games, there may be entities other than cards in it. \looseness=-1

Games may also have additional zones. For instance, \fab, \hs, and \yugioh have a \textbf{secrets} zone, in which cards are not visible to opponents. \tot has a shared zone: the \textbf{tavern}. Finally, in \pokemon, each player has a \textbf{prizes} zone, filled at the start of the battle with a number of face-down cards drawn from the top of the player's deck.

Most CCGs zones are \textbf{unordered}: card order on the zone is irrelevant. We call other zones \textbf{positional}. 
In all CCGs, players' decks are positional because the order in which cards are drawn is key. Some CCGs like \gwt and \hs, some effects consider creatures ``around'', ``in front of'', and other positional indicators, so their battlefield and lanes are positional. \yugioh also has creature summoning mechanics in which position on the battlefield matters.

Zones also differ in visibility. We identify three types: size, content, and order. Generally, all players are allowed to know \textit{how many cards} are in any given zone (the zone's \textbf{size}).
Some zones restrict the opponent from knowing \textit{which cards} are in them (the zone's \textbf{content}). Lastly, in positional zones, knowledge of the cards' specific \textbf{order} may be unavailable. The deck zone is an illustrative example: all players know its size, its owner knows its content, and no one knows its order. Table \ref{tab:zones} specifies the visibility of the most common zones.\looseness=-1 

\begin{table}[htbp]
    \setlength\tabcolsep{0.5em}
    \centering
    \caption{Zones, their types and visibilities.}
    \label{tab:zones}
    \begin{tabular}{lcccc}
        \toprule
        \multirow{2}{*}{\textbf{Zone}} & \multirow{2}{*}{\textbf{Type}} & \multicolumn{3}{c}{\textbf{Visibility}} \\\cmidrule[\heavyrulewidth]{3-5}
        & & \textbf{Size} & \textbf{Content} & \textbf{Order} \\\midrule[\heavyrulewidth]
        Deck & Positional & Everyone & Owner & No one \\\midrule
        Hand & Unordered & Everyone & Owner & - \\\midrule
        Battlefield or Lane & May be both & Everyone & Everyone* & Everyone \\\midrule
        Stack or Queue & Positional & Everyone & Everyone & Everyone \\\midrule
        Graveyard & May be both & Everyone & Everyone & Everyone \\\midrule
        Outside the game & Unordered & Everyone & Everyone & - 
        \\\bottomrule
    \end{tabular}
    \\
    \vspace{0.25em}
    \raggedright\footnotesize{* In \yugioh, creature cards may be put face down on the battlefield. In this case, its content visibility would be ``owner''.}
\end{table}

\subsection{Combat}
\label{sec:combat}

The combat in combat-based CCGs can be classified as declared or targeted. In CCGs with \textbf{declared combat}, a player declares the intent of attacking with a subset of their creatures, and then the opponent may decide to block any attacking creature with their own creatures. Then, attacking creatures and their respective blocking creatures damage each other, and any unblocked attacks damage the opponent. In \textbf{targeted combat}, players declare attack one creature at a time, and attacks have a target---a specific opponent's creature or the opponent themselves. Any damage due is dealt immediately after declaring the attack, with no interaction with the opponent. 

Damage done to the opponent subtracts from their health points, while damage done to creatures subtracts from their health attribute. When a creature's health becomes zero or negative, it is destroyed and goes to the graveyard zone, if applicable. From our list, \mtg is the only game where creatures recover from damage, returning to their original health at the end of the turn. Creatures commonly possess combat-related abilities that alter combat rules. 

\subsection{Actions}
\label{sec:actions}

We have described many moments where players choose among a set of available actions. Now, we list these actions and position them temporally in gameplay.

Every gameplay of a CCG involves deck-building and battling. During constructed deck-building, the player may first be prompted to \textbf{select a deck class} (in a hard-classed CCG) and then to \textbf{build a deck}. 
In a draft-based deck-building process, the player passes through many turns of draft, in which they \textbf{pick a card} among several alternatives. Then, picked cards either directly become the player's deck or serve as a card pool from which to build a deck. 

The battle is the most complex part of gameplay in terms of actions. In summary, as the battle starts, the player may be offered a chance to \textbf{redraw} their entire hand or specific cards in their hand. Then, players start taking turns, and the player will be prompted to act---be it in a main phase, a combat, or in response to an opponent's action. At this moment, the player must select one of the available actions, that may include \textbf{playing a card}, \textbf{activating an ability}, \textbf{attacking}, \textbf{blocking}, or doing nothing (\textbf{no-op}). Some actions may also require, as a consequence, that the player decides how to \textbf{spend their mana}. Not all actions are available at all turn phases, as Table \ref{tab:action-validity} explains, and even when they are, some combinations of their parameters may not be valid (for instance, playing a card that costs more than the available mana or attacking with a creature that cannot attack). 

After performing an action, the battle continues until the player is again prompted to act. This repeats indefinitely until a win condition is reached. 

\begin{table}[tbp]
    \setlength\tabcolsep{0.5em}
    \centering
    \caption{Player actions in a battle and their validity in each turn phase.}
    \label{tab:action-validity}
    \begin{tabular}{lcccc}
        \toprule
        \multirow{2}{*}{\textbf{Action}} & \multicolumn{4}{c}{\textbf{Availability}} \\\cmidrule{2-5}
         & \textbf{Main} & \textbf{Combat} & \textbf{Response} & \textbf{End} \\\midrule
        Card play & Yes & Fast cards only & Fast cards only & No** \\\midrule
        Ability activation & Yes & Yes & Yes & No** \\\midrule
        Attack & No* & Yes & No & No \\\midrule
        Block & No* & Yes & No & No 
        \\\bottomrule
    \end{tabular}
    \\
    \vspace{0.25em}
    \raggedright\footnotesize{* In CCGs with no combat phase, attacking and blocking occur on the main phase.}
    \\
    \raggedright\footnotesize{** In \mtg, fast cards can be played, and abilities can be activated during the end phase.}
\end{table}

\subsection{Win Conditions}
\label{sec:win-con}

The typical win condition of CCGs is reducing the opponent's health points to zero (or fewer). To achieve this, players typically play and use creatures in combat. 
Spells and other cards may help by doing damage directly or by facilitating creatures to do so. We call games that follow this archetype \textbf{combat-based CCGs}. 
This includes most games in our study. 
We additionally consider \pokemon as such 
despite not having 
health points, but awarding the win to the first player to empty their prize zone (players draw from there every time they defeat an opponent's creature).

\gwt, \ms, and \tot compose a second archetype of CCGs that, instead, require the player to accumulate \textit{influence} points. 
In these \textbf{conquest-based CCGs}, creatures 
and some spells provide points, and at the end of the battle, the player with the most points wins. Consequently, 
turns are limited either directly (6 turns in \ms) or indirectly 
(running out of cards in \gwt or reaching 40 points in \tot). While there is no combat, there are means to remove or weaken creatures. 
\looseness=-1

\tot also implements an unique win condition. At the start
, each player chooses two patrons, each with distinct activated abilities and starting in a neutral stance. Whenever a player uses a patron's ability, that patron starts favoring that player if it is neutral or becomes neutral if it favors the opponent. Gaining the favor of all four patrons results in immediate victory.

Running out of cards to draw in CCGs is usually penalized with an immediate loss. 
Lastly, we point out that CCGs have cards that introduce novel win conditions. For instance, with ``Felidar Sovereign'' from \mtg, one wins by having 40 or more health points, and in \yugioh, by having in their hand ``Exodia the Forbidden One'' and its other four component cards.

\section{Discussion}
\label{sec:discussion}

The stated goals of this work regarding CCGs were to analyze the differences and similarities among rules, mechanics, and game modes, which we addressed with a taxonomy of CCGs. 
Despite our perceived heterogeneity across the genre
, there was sufficient common structure among the games for such taxonomy to be possible. 
In the next paragraphs, we discuss some aspects of our work.

\textbf{Game complexity}.
Across the games we selected, we observed that the number of taxonomy elements (e.g., resource system, combat, etc) serves as a proxy for the complexity of each game. For instance, \locm, a simplified game made for AI research, and \ms, a fast-paced mobile game aimed at a broader audience, exhibit the fewest number of elements. 
Meanwhile, \mtg, one of the most complex games known, and \fab, heavily inspired by it, display the opposite pattern. 

Another observation regarding complexity was that some action types may benefit from a computational complexity analysis. 
For example, deciding whether there is a combination of mana-generating abilities that satisfies a (complex) mana cost or blocking a declarative attack in a way 
that maximizes the player's health points may be NP-hard or harder. Also, some action types resemble subset selection problems, which are known to be NP-hard \cite{BinshtokBSMB07}. As discussed in Section \ref{sec:rel-work}, CCGs have been a frequent target of such studies.

\textbf{Design of AI players}.
AI players can analyze multiple scenarios much faster than humans and can perfectly recall past moves. 
Despite this, a human with knowledge of the game could address all action types discussed in Section \ref{sec:actions}, whereas most AI techniques would require some degree of compartmentalization 
(e.g., different models for each action type). As an approach moves towards considering all action types in a unified manner, it starts to resemble a general decision maker. In fact, CCGs have been compared to a general game-playing task in the past \cite{ChoeK19,CowlingWP12}.

\textbf{Limitations}. Our approach is just one of many others that could be proposed. However, it is one that we believe is representative of the genre and may be helpful for future works. 
Still, the resulting analysis is not exhaustive, and some specific aspects of CCGs might not have been included.

\section{Conclusion}
\label{sec:conclusion}

We proposed a taxonomy of collectible card games by studying their rule books, \textit{wikis}, and other sources. 
We categorized CCGs by win condition, combat mechanism, resource systems, and other aspects relevant to game-playing AI. 
To the best of our knowledge, this is the first work proposing such a taxonomy. 
We provided a novel perspective on the genre of CCGs: 
one that characterizes them under a single framework while encompassing their complexity and heterogeneity. We believe it will benefit and foster discussion among researchers, game designers, and enthusiasts in understanding the games and the field of CCG AI.
As future work, we aim to formally define the many decision problems players face during CCG matches, as introduced in Section \ref{sec:actions}.

\begin{credits}
\subsubsection{\ackname} This work was partially supported by CAPES (finance code 001), CNPq, and Fapemig.

\subsubsection{\discintname}
The authors have no competing interests to declare that are relevant to the content of this article.
\end{credits}

\bibliographystyle{splncs04}
\bibliography{refs}

\end{document}